\documentclass[12pt,a4paper]{article}
\usepackage{amsthm,amsmath,amsfonts,amssymb}
\usepackage{graphicx}
\usepackage{multirow}
\usepackage{mathrsfs}
\usepackage{xcolor}
\usepackage{url}
\usepackage[final]{microtype}
\usepackage[authoryear,round]{natbib}
\usepackage[colorlinks,citecolor=blue,urlcolor=blue]{hyperref}

\usepackage[a4paper,textwidth=42pc,textheight=658pt,centering,includeheadfoot]{geometry}
\theoremstyle{plain}

\theoremstyle{definition}

\usepackage{booktabs}
\usepackage{algorithm,ulem}
\usepackage{algpseudocode}
\usepackage{bbm}
\usepackage{bm}
\usepackage{comment}

\usepackage{titlesec}
\titleformat{\subsubsection}[block]{\normalfont\normalsize\itshape}{\thesubsubsection}{1em}{}

\graphicspath{{./figures}}

\usepackage{enumitem}

\newcommand{\bc}{\begin{cases}}
	\newcommand{\ec}{\end{cases}}

\DeclareMathOperator*{\argmin}{argmin}
\DeclareMathOperator*{\argmax}{argmax}

\newcommand{\bbR}{{\mathbb R}}

\newcommand{\bv}{{\boldsymbol{v}}}
\newcommand{\bw}{{\boldsymbol{w}}}

\newcommand{\bepsilon}{{\boldsymbol{\varepsilon}}}

\newcommand{\bx}{{\boldsymbol{x}}}
\newcommand{\bX}{{\boldsymbol{X}}}
\newcommand{\bY}{{\boldsymbol{Y}}}
\newcommand{\by}{{\boldsymbol{y}}}
\newcommand{\bZ}{{\boldsymbol{Z}}}
\newcommand{\bz}{{\boldsymbol{z}}}

\newcommand{\E}{\mathbb{E}}

\newcommand{\CF}{\mathcal{F}}

\newcommand{\bbN}{\mathbb{N}}

\newcommand{\wbar}[1]{\overline{#1}}

\definecolor{darkgreen}{rgb}{0,0.6,0}

\def\hat{\widehat}

\makeatletter
\renewcommand\section{\@startsection{section}{1}{\z@}%
  {-3.5ex \@plus -1ex \@minus -.2ex}%
  {2.3ex \@plus .2ex}%
  {\normalfont\Large\bfseries\raggedright}}
\renewcommand\subsection{\@startsection{subsection}{2}{\z@}%
  {-3.25ex\@plus -1ex \@minus -.2ex}%
  {1.5ex \@plus .2ex}%
  {\normalfont\large\bfseries\raggedright}}
\makeatother

\allowdisplaybreaks
\begin{document}

\title{\Large\textbf{Generative and Nonparametric Approaches for Conditional Distribution Estimation: Methods, Perspectives, and Comparative Evaluations}}
 
\author{\normalsize Yen-Shiu Chin$^1$ , Zhi-Yu Jou$^1$, Toshinari Morimoto$^2$, Chia-Tse Wang$^3$,\\ \normalsize  Ming-Chung Chang$^1$, Tso-Jung Yen$^1$, Su-Yun Huang$^1$, and Tailen Hsing$^4$ \\[1ex]
\small $^1$Institute of Statistical Science,
Academia Sinica, Taipei, Taiwan\\
\small $^2$Department of Mathematics,
National Taiwan University, Taipei, Taiwan\\
\small $^3$Data Science Degree Program, National Taiwan University and Academia Sinica, Taipei, Taiwan\\
\small $^4$Department of Statistics,
University of Michigan, Ann Arbor, USA
}

\date{}
\maketitle

\begin{abstract}
The inference of conditional distributions
is a fundamental problem in statistics, essential for prediction, uncertainty quantification, and probabilistic modeling. A wide range of methodologies have been developed for this task. This article reviews and compares several representative approaches spanning classical nonparametric methods and modern generative models. We begin with the single-index method of \cite{hall2005approximating}, which estimates the conditional distribution through a dimension-reducing index and nonparametric smoothing of the resulting one-dimensional cumulative conditional distribution function. We then examine the basis-expansion approaches, including FlexCode \citep{izbicki2017converting} and DeepCDE \citep{dalmasso2020conditional}, which convert conditional density estimation into a set of nonparametric regression problems. In addition, we discuss two recent generative simulation-based methods that leverage modern deep generative architectures: the generative conditional distribution sampler \citep{zhou2023deep} and the conditional denoising diffusion probabilistic model \citep{fu2024unveil,
yang2025conditional}. A systematic numerical comparison of these approaches is provided using a unified evaluation framework that ensures fairness and reproducibility. The performance metrics used for the estimated conditional distribution include the mean-squared errors of conditional mean and standard deviation, as well as the Wasserstein distance. We also discuss their flexibility and computational costs, highlighting the distinct advantages and limitations of each approach.\\

\noindent \textbf{Key words and phrases:} Conditional distribution estimation, Deep learning, Diffusion models, Dimension reduction, Generative modeling, Wasserstein distance.
\end{abstract}

\twocolumn  

\section{Introduction}
The inference of conditional distributions and related functionals (e.g., mean, quantiles, variance, etc.) is a fundamental statistical task underlying  prediction, uncertainty quantification, and probabilistic modeling.
Numerous classical and modern methods exist for the general problem of conditional distribution inference. However, comparisons across the diverse approaches, particularly between the more traditional nonparametric and the newer generative simulation-based methods remain limited.
This article reviews some representative approaches and provides a systematic comparison.

Let $(\bX,\bY)$ denote a random pair with $\bX\in\mathcal{X} \subseteq\bbR^p$ and $\bY\in\mathcal{Y} \subseteq\bbR^q$ for $p,\;q \geq 1$. We assume that the joint distribution of $(\bX,\bY)$ is $F_{\bX,\bY}$, and denote the corresponding marginal distributions by $F_\bX$ and $F_\bY$. Let $F_{\bY\mid \bX}(\cdot\mid \bx)$ represent the distribution of $\bY$ conditional on $\bX=\bx$, and $f_{\bY\mid \bX}(\cdot\mid \bx)$ denote its corresponding density. We shall use the non-boldface notation $Y$ in place of $\bY$ when referring explicitly to the  univariate case $\mathcal{Y}\subseteq \bbR$.

A key theoretical foundation for conditional distribution estimation is the so-called Noise-Outsourcing Lemma (see, e.g., \citet[Lemma 4.22 and Chapter 8]{kallenberg2021foundations}, \citet[Lemma 3.1]{Austin15}), which states that if $\mathcal{X}$ and $\mathcal{Y}$ are standard Borel spaces, then on an enlarged probability space there exist a random variable $U\sim\mbox{Uniform}[0,1)$ and a Borel-measurable function $G: [0, 1) \times \mathcal{X} \to\mathcal{Y}$, where $U$ and $\bX$ are independent, such that
\begin{align} \label{e:outsourcing}
\bY= G(U,\bX) \mbox{ a.s.}
\end{align}
Thus, the conditional distribution of $\bY\mid \bX=\bx$ can be represented as the distribution of $G(U,\bx)$, where $U$ encodes the variability in $\bY$ that remains unexplained by $\bX$. Note, however, that both $G$ and $U$ are generally non-unique.

This lemma also reveals a close connection to the classical nonparametric regression model:
\begin{align} 
\label{e:reg}
\bY= g(\bX) + \boldsymbol{\varepsilon},
\end{align}
where randomness beyond the conditional mean is absorbed by a zero-mean noise term $\boldsymbol{\varepsilon}$, and the main goal is to infer $g(\bx)=\E(\bY\mid \bX=\bx)$.
Although the study of \eqref{e:reg} occupies a central place in the statistical literature, our focus here is not on nonparametric regression itself. Since the conditional mean is a statistical functional of the conditional distribution, a conditional distribution estimation method can, in principle, be adapted to conditional mean estimation; see FlexCode \citep{izbicki2017converting} and DeepCDE \citep{dalmasso2020conditional} in Section~\ref{s:approaches} for examples on this for the univariate case $Y\in\bbR$. Importantly, the representation~\eqref{e:outsourcing} provides the theoretical underpinning for generative simulation-based approaches, which learn the mapping $G(U, \bX)$ directly to model the conditional distribution of $\bY \mid \bX$. These ideas motivate the two generative methods reviewed in Section~\ref{s:approaches}.

The traditional literature on the conditional distribution inference problem primarily focused on settings with small $p$ and $q$. 
\cite{hall1999methods} considered an adjusted Nadaraya-Watson estimator of the conditional cumulative distribution function. \cite{yu1998local} and \cite{spokoiny2013local} considered the inference of the quantile function of the conditional distribution.
\cite{dalmasso2020conditional} introduced several approaches on the estimation of conditional density, including a nearest-neighbor-based kernel density estimator and an approach based on basis expansion that turns the estimation of the conditional density into a series of nonparametric regression problems. These approaches are kernel based and as such are sensitive to the ambient dimensions of both $\mathcal{X}$ and $\mathcal{Y}$.

Some recent studies have developed approaches that are more flexible with respect to $p$, the dimension of $\bX$.
For example, \cite{hall2005approximating} introduced a dimension-reduction approach for models where the conditional distribution of a scalar random variable $Y$ given $\bX=\bx$ depends approximately on a single index $\bv^\top \bx$ for some direction $\bv$, thereby potentially overcoming some of the challenges posed by large $p$. This idea has the same flavor as some classical dimension-reduction strategies in regression; see also \cite{henzi2023distributional} for a related approach that combines the ideas of single-index models and isotonic regression. The basis-expansion and nonparametric regression approach of \cite{dalmasso2020conditional} does not directly address the curse of dimensionality due to large $p$. However, their deep-neural-network implementation (\texttt{DeepCDE}) 
may achieve a similar effect, as deep networks can implicitly learn lower-dimensional representations of the predictor space \citep[see, e.g.,][]{hinton2006reducing,bauer2019,schmidt2020nonparametric}. 

Recently, two generative simulation-based approaches were proposed in \cite{zhou2023deep}, and \cite{fu2024unveil} and \cite{yang2025conditional}. Both methods take advantage of recent advances in deep neural networks. The method of \cite{zhou2023deep}, known as the Generative Conditional Distribution Sampler (GCDS), estimates the mapping $G$ in~\eqref{e:outsourcing} by training a neural network $\widehat G$ that minimizes the empirical Kullback–Leibler divergence between a candidate joint distribution and the observed data distribution. Then, for a given $\bx$, the conditional distribution of $\bY\mid \bX=\bx$ is simulated by generating samples of a latent noise variable (e.g., a multivariate Gaussian vector $\bm\eta$; see Section~\ref{s:GCDS}) and evaluating $\widehat G(\bm{\eta},\bx)$. In contrast, \cite{fu2024unveil} and \cite{yang2025conditional} considered
the simulation of the conditional distribution using the conditional diffusion model, which is a conditional extension of the denoising diffusion probabilistic model (DDPM; \citealp{ho2020denoising}) and will be referred to as the conditional DDPM.
Both of these offer greater flexibility with respect to the dimensions $p$ and $q$ than kernel-based approaches
and are capable of efficient estimation of various functionals of the conditional distribution.

In this article, we review and provide a systematic numerical comparison of four representative approaches for conditional distribution estimation: (i) the single-index model approach \citep{hall2005approximating}, (ii) basis-expansion methods including FlexCode \citep{izbicki2017converting} and DeepCDE \citep{dalmasso2020conditional}, (iii) the distribution-matching generative model GCDS \citep{zhou2023deep}, and (iv) the conditional diffusion-based generative model \citep{fu2024unveil, yang2025conditional}. All four approaches can potentially be applied in the setting of high-dimensional predictors.
The approaches (i) and (ii) are selected for their superior flexibility and effectiveness among kernel and related nonparametric estimation methods. 
For (iii) and (iv), since a substantial body of research has already examined how well generative models perform in ultra high-dimensional data contexts, our aim here is to study their performance in scenarios where the dimensionality is more moderate and kernel-based methods have been commonly used.

The remainder of the article is organized as follows. In Section~\ref{s:approaches}, we briefly review the four representative approaches, (i)-(iv), mentioned above, 
In Section~\ref{s:simulations}, we conduct a comprehensive set of simulation studies to compare the effectiveness of these methods. Section~\ref{s:discussion} offers additional discussion and concluding remarks.

To facilitate systematic evaluations and ensure reproducibility, we developed a unified Python framework that integrates our implementations with the original software of the authors when available, including packages previously released only in \textsf{R} or not publicly released. 
The full codebase is publicly accessible on GitHub.\footnote{\url{https://github.com/chiatsewang/generative-nonparametric-cde}}

\section{A review of four classes of methods} 
\label{s:approaches}

This section reviews several methods for estimating conditional distributions or densities, whose empirical performance will be compared in the simulation studies of Section~\ref{s:simulations}. The approaches considered here include those introduced in \cite{hall2005approximating}, \cite{izbicki2017converting},
\cite{dalmasso2020conditional}, \cite{zhou2023deep},
 and \cite{fu2024unveil}, and \cite{yang2025conditional}. One of the criteria for selecting these methods is their potential to accommodate high-dimensional predictors.

\subsection{Hall and Yao: a dimension-reduction approach}

We begin with the dimension-reduction approach of \cite{hall2005approximating}, developed for scalar responses
$Y\in\bbR$ with predictors $\bX\in \bbR^p$. The basic idea is to approximate the conditional cumulative distribution function $F_{Y\mid \bX}(y\mid \bx)$ by a single-index model of the form
$$F(y\mid \bv^\top \bx) := P(Y\le y\mid \bv^\top \bX = \bv^\top \bx)
$$ for some unit vector $\bv$ that is not known a priori. This unknown $\bv$ represents the direction in the predictor space that captures the dominant dependence of $Y$ on $ \bX$. For certain models, this approximation is exact, while for others it provides a useful low-dimensional approximation of the conditional distribution.  
A key question is how to estimate the direction vector $\bv$. \cite{hall2005approximating} proposed  minimizing a discrepancy measure between the true conditional probability and its single-index approximation, briefly explained below.

Let $A$ be a Borel set of $\bbR^p$ and define
$$
F({A}, y)=P\{( \bX, Y) \in {A} \times(-\infty, y]\};
$$
also, let
$$
H_\bv({A}, y)=\int_{A} F\left(y \mid \bv^\top \bx\right) dF_{\bX}(\bx).
$$
If for some $\bv$ and all $\bx$, the single-index approximation $F(y\mid\bv^\top \bx)\approx F_{Y\mid \bX}(y\mid\bx)$ holds well, then $F({A}, y)$ and $H_\bv({A}, y)$ should be close for a rich class of ${A}$'s. \cite{hall2005approximating} focused on the class 
$Q=\left\{\boldsymbol{\delta}: {A}_{\boldsymbol{\delta}} \subseteq \mathscr{R}\right\}$,
where each $A_{\boldsymbol{\delta}}$ is a $p$-dimensional sphere contained within a given fixed set $\mathscr{R} \subseteq \mathbb{R}^p$, indexed by a $(p+1)$-vector
$\boldsymbol{\delta}$, whose first $p$ components denote the center and whose last component specifies the radius.

Suppose we know $F_Y(y), F({A}, y)$, and $H_\bv({A}, y)$.
We can compute the optimal $\bv$ by minimizing the following criterion over $\bv \in \Theta$, where $\Theta$ denotes the set of $p$-dimensional unit vectors whose first nonzero component is positive:
\begin{align} \label{e:S_0}
S_0(\bv)=\int_Q \int_{-\infty}^\infty\left\{F\left({A}_{\boldsymbol{\delta}}, y\right)-H_\bv\left({A}_{\boldsymbol{\delta}}, y\right)\right\}^2 dF_Y(y)\, d\boldsymbol{\delta}.
\end{align}
However, in general, the functions in \eqref{e:S_0} are unknown. In practice, $S_0(\bv)$ is approximated by
$$
S(\bv) =\int_Q \ S\left(\bv, {A}_{\boldsymbol{\delta}}\right) d\boldsymbol{\delta}, 
$$
where $S(\bv,A)$ is defined, based on the observed data $(\bX_i, Y_i)$, $i=1,\dots,n$, as
\begin{align*}
S(\bv, {A}) & := {1\over n} \sum_{j=1}^n\Bigg\{\widehat{F}_{-j}\left({A}, Y_j\right)\\
&\hspace{.7cm} -\frac{1}{n-1} \sum_{i: i \neq j, \bX_i \in {A}} \widehat{F}_{-i,-j}\left(Y_j \mid \bv^\top \bX_i\right)\Bigg\}^2.
\end{align*}
Here, $\widehat{F}_{-j}(A,Y_j)$ is the empirical estimate of $F(A,Y_j)$ based on all data except $(\bX_j,Y_j)$, and
$\widehat{F}_{-i,-j}(Y_j \mid \bv^\top \bX_i)$ is a leave-two-out local linear estimator of $F(Y_j\mid \bv^\top \bX_i)$.
Thus, $S(\bv,A_{\boldsymbol{\delta}})$ provides an estimate of 
$$
\int_{-\infty}^\infty\left\{F\left({A}_{\boldsymbol{\delta}}, y\right)-H_\bv\left({A}_{\boldsymbol{\delta}}, y\right)\right\}^2 dF_Y(y).
$$
We then define $\hat{\bv}$ as the minimizer of $S(\bv)$ over $\bv \in \Theta$.

Finally, the conditional cumulative distribution function $F_{Y\mid \bX}(y\mid \bx)$ is estimated by $\widehat{F}\left(y \mid \hat{\bv}^\top \bx\right)$, which is a local linear estimator of $F\left(y \mid \hat{\bv}^\top \bx\right)$ and constructed in the same way as $\widehat{F}_{-i,-j}$ but using the full sample (without leaving two out). 
Different bandwidths may be employed for estimating $\bv$ and $F\left(y \mid \hat{\bv}^\top \bx\right)$.
\cite{hall2005approximating} provided an empirical rule for choosing the bandwidths and developed a convergence rate for the estimated conditional cumulative distribution function under suitable conditions. Some advantages and disadvantages of the Hall and Yao approach are as follows.

\vskip.3cm
Advantages:
\begin{itemize}
\item 
The method mitigates the curse of dimensionality by reducing the predictor space from $\bbR^p$ to one dimension through a single index model. 
\item 
A rigorous asymptotic theory is available.
\end{itemize}

Disadvantages:
\begin{itemize}
    \item Single-index limitation: The approximation of \newline $F_{Y\mid \bX}(y\mid \bx)$ by $F(y\mid \bv^\top \bx)$ may suffer from model misspecification. Extensions to multiple-index settings are nontrivial, and to our knowledge no such extensions have been developed.
    \item 
    Multi-dimensional $\bY$: It seems possible to consider the case where $\bY$ is a vector by focusing on its joint cdf. However, the implementation in that case will undoubtedly add an extra layer of challenges.
    \item Computational challenges: Estimating $\bv$ involves minimizing a nonconvex, high-dimensional objective function based on kernel estimators of conditional distributions, which is computational demanding.
\end{itemize}

\subsection{\texttt{FlexCode} and \texttt{DeepCDE}: nonparametric estimation using orthogonal series expansion}

Next, we review the basis-expansion approaches of \cite{izbicki2017converting} and
\cite{dalmasso2020conditional}, which considered the estimation of the conditional density $f_{Y\mid \bX}(y\mid \bx)$ of a scalar response $Y\in\bbR$ given a random predictor $\bX\in\bbR^p$. Let $\{\varphi_j\}_{j=1}^\infty$ be an orthonormal basis in $L^2(\bbR)$, such as a cosine, Fourier or wavelet basis. Then, we can write
\begin{align} \label{np_reg}
f_{Y\mid \bX}(y\mid \bx) & = \sum_{j=1}^\infty \beta_j(\bx)\varphi_j(y), 
\end{align}
where the coefficients are
\begin{align*} \label{np_reg_1}
\begin{split}
\beta_j(\bx) &= \int f_{Y\mid \bX}(y|\bx)\varphi_j(y)dy \\ 
& = \E\left(\varphi_j(Y)\mid \bX=\bx\right).
\end{split}
\end{align*}
Thus, the problem of estimating $f(y\mid \bx)$ becomes a nonparametric regression problem of estimating the functions $\beta_j(\bx)$. This can, in principle, be accomplished using any suitable nonparametric regression method applied to the data pair $\{\bX_i,\varphi_j(Y_i), i=1,\ldots,n\}$. In practice, the basis expansion must be truncated, with the number of retained terms serving as a tuning parameter that balances bias and variance in the final density estimate. 
\cite{izbicki2017converting} considered several nonparametric regression techniques for estimating the $\beta_j(\bx)$, depending on the data type, including sparse additive models, nearest-neighbor regression, random forests, and support distribution machines. 
\cite{dalmasso2020conditional} further extended this framework by employing neural networks for the estimation of $\beta_j(\bx)$. The FlexCode approach has implementations in both \textsf{R} (\url{https://github.com/rizbicki/FlexCoDE}) and \textsf{Python} (\url{https://github.com/lee-group-cmu/FlexCode}), and DeepCDE is available in \textsf{Python} at \url{https://github.com/lee-group-cmu/DeepCDE/tree/master}.

\vskip.3cm
Advantages: 
\begin{itemize}
\item 
The formulation of expressing $f_{Y\mid \bX}(\cdot \mid \bx)$ by a basis expansion is conceptually clean.
The coefficients estimation can be regarded as nonparametric regression problems, which can be solved using a wide range of methods, allowing flexibility and adaptability to different data types and data structures. 
\item 
Training for FlexCode is computationally efficient. Moreover, although parallelization is not essential in typical problem sizes, the method is naturally parallelizable over basis indices $j$ because since each regression problem can be handled independently.
\end{itemize}

Disadvantages: 
\begin{itemize}
\item
FlexCode and DeepCDE rely on the assumption that the conditional density admits a well-behaved orthogonal series representation of the form~\eqref{np_reg}, so that a finite number of basis coefficients can appropriately capture the relevant structure of $f_{Y\mid \bX}$. 
In more complex scenarios, for example those involving heteroscedastic noise—such as \ref{M6:heteroscedastic_noise} and \ref{M7:heteroscedastic_covariate-dependent_mixture_noise} in Section~\ref{s:simulations}—the performance of FlexCode and DeepCDE tends to deteriorate.

\item 
The estimated conditional density may take negative values or fail to integrate to one. Additional steps, such as post-processing or normalization, are often required to ensure that the estimate satisfies the basic properties of a valid density function.
\item
The series expansion must be truncated at some finite $J$, and the performance of the approach depends on this choice. 
The cutoff $J$ serves as a tuning parameter that balances the bias-variance tradeoff in the resulting density estimator. Typically, smoother densities can be adequately approximated with a smaller $J$.
\end{itemize}

\subsection{\texttt{GCDS}: a distribution matching generative approach}
\label{s:GCDS}

The third approach we consider is proposed by~\cite{zhou2023deep}, which is fundamentally different from the previous two methods. It is a simulation-based generative approach grounded in the Noise Outsourcing Lemma introduced in~\eqref{e:outsourcing}. In this framework, the uniform noise variable $U$ is replaced by a latent noise vector $\boldsymbol{\eta}\sim N\left(\mathbf{0}, \mathbf{I}_m\right)$ that is independent of $\bX$.
The vector $\boldsymbol{\eta}$ introduces stochasticity into the generative mechanism and represents the randomness in $\bY$ not explained by $\bX$. Although the statement of the Noise Outsourcing Lemma uses a scalar uniform variable, it is common in practice to employ a multivariate Gaussian noise vector, which provides richer latent variability and enables the model to capture complex or multimodal conditional distributions.
The latent dimension $m$ governs the expressive capacity of the model: a larger $m$ allows more flexible approximations but may also lead to greater stochastic variation. It is worth noting that $m$ need not equal $q$, the dimension of $\bY$. Intuitively, GCDS trains a generator to produce synthetic pairs $(\bX, G(\boldsymbol{\eta}, \bX))$ whose joint distribution matches the observed joint distribution of $(\bX, \bY)$.

In the approach of \cite{zhou2023deep}, the measurable mapping $G$ is parameterized by a neural network $G_\psi$ with parameters $\psi$. 
Following the ideas of generative adversarial networks \citep{goodfellow2014generative}, \cite{zhou2023deep} adopted a distribution-matching approach that estimates the conditional generator $G_\psi$ by minimizing the Kullback-Leibler (KL) divergence, $\mathbb{D}_{\mathrm{KL}}\left(F_{\bX, G_\psi} \| F_{\bX, \bY}\right)$, between the joint distribution of
the generated pair, $(\bX,$ $G_\psi(\boldsymbol{\eta},\bX))$, and the true joint distribution of $(\bX,\bY)$.
Let
\begin{equation}\label{eq:log_ratio}
D(\bw) = \log \frac{f_{\bX,G_\psi}(\bw)}{f_{\bX,\bY}(\bw)},\quad \bw\in\bbR^{p+q},
\end{equation}
which is the logarithm of the corresponding density ratio.
Here $D(\bw)$ denotes the ideal log-density ratio in~\eqref{eq:log_ratio}, while $D_\phi(\bw)$ is its neural-network approximation parameterized by~$\phi$.
Using the variational representation~\citep{nguyen2010estimating}, the objective KL divergence can be reformulated 
as a minimax optimization problem over a generator $G_\psi$ and a discriminator $D_\phi$. Thus,
\begin{align*}
& \mathbb{D}_{\mathrm{KL}}\left(F_{\bX, G_\psi} \| F_{\bX, \bY}\right) \\
&= \E_{(\bX,\boldsymbol{\eta})}\!\left[
\log \frac{f_{\bX,G_\psi}(\bX,G_\psi(\boldsymbol{\eta},\bX))}{f_{\bX,\bY}(\bX,G_\psi(\boldsymbol{\eta},\bX))}
\right]\\
&= \sup _{D_\phi}\Big\{\mathbb{E}_{(\bX, \boldsymbol{\eta}) \sim F_\bX F_{\boldsymbol{\eta}}}[D_\phi(\bX, G_\psi(\boldsymbol{\eta}, \bX))] \\
& \hspace{1cm} -\mathbb{E}_{(\bX, \bY) \sim F_{\bX, \bY}}[\exp (D_\phi(\bX, \bY)-1)]\Big\} \\
&= \sup _{D_\phi}\Big\{\mathbb{E}_{(\bX, \boldsymbol{\eta}) \sim F_\bX F_{\boldsymbol{\eta}}}[D_\phi(\bX, G_\psi(\boldsymbol{\eta}, \bX))] \\
& \hspace{1cm} -\mathbb{E}_{(\bX, \bY) \sim F_{\bX, \bY}}[\exp (D_\phi(\bX, \bY))]\Big\}+1,
\end{align*}
where $F_{\boldsymbol{\eta}}$ denotes the distribution function of $\boldsymbol{\eta}$.
Define
\begin{align*}
\mathcal{L}(G_\psi, D_\phi)
= &\;\mathbb{E}_{(\bX, \boldsymbol{\eta}) \sim F_\bX F_{\boldsymbol{\eta}}}[D_\phi(\bX, G_\psi(\boldsymbol{\eta}, \bX))] \\
&-\mathbb{E}_{(\bX, \bY) \sim F_{\bX, \bY}}[\exp (D_\phi(\bX, \bY))].
\end{align*}
Thus, $G_\psi$ and $D_\phi$ are estimated by minimizing a sample version of $\mathcal{L}(G_\psi, D_\phi)$. Given an i.i.d.\ sample
$\{(\bX_i,\bY_i),i= 1,\ldots,n\}$ from $F_{\bX,\bY}$ and an independent sample $\{\boldsymbol{\eta}_i,i=1,\ldots,n\}$ generated from $F_{\boldsymbol{\eta}}$, define
\[
\begin{split}
&\widehat{\cal{L}}(G_\psi,D_\phi)\\
=& {1\over n} 
\sum_{i=1}^n 
\left\{D_\phi(\bX_i,G_\psi(\boldsymbol{\eta}_i,\bX_i))-
\exp(D_\phi(\bX_i,\bY_i))\right\}.
\end{split}
\]
The first term in $\widehat{\mathcal{L}}(G_\psi, D_\phi)$ uses samples
$(\bX_i, \boldsymbol{\eta}_i)$ drawn from $F_\bX \times F_{\boldsymbol{\eta}}$, where the noise variables
$\boldsymbol{\eta}_i$ are independently re-drawn at each training epoch. The
second term uses the observed pairs $(\bX_i, \bY_i) \sim F_{\bX,\bY}$ to
approximate the expectation under $F_{\bX,\bY}$.
Here, $G_\psi, D_\phi$ are two separate feedforward neural networks \citep{Goodfellow-et-al-2016}. For network parameter selection, we solve
\begin{align*}
    (\hat\psi,\, \hat\phi) =\argmin_\psi\argmax_\phi \widehat{\cal{L}}(G_\psi,D_\phi),
\end{align*}
and set $\widehat G = G_{\hat\psi}$. As shown in~\cite{zhou2023deep}, under suitable conditions, the distribution of $\widehat G(\boldsymbol{\eta},\bx)$ provides a consistent estimator of the conditional distribution of $\bY$ given $\bX = \bx$.

\vskip.3cm
Advantages: 
\begin{itemize}
\item
Sampling during inference is speedy. Once the model is trained, generating samples requires only a single forward pass of the generator, making it much faster at inference time than diffusion-based methods such as DDPM, which involve multiple iterative denoising steps.
\end{itemize}

Disadvantages:
\begin{itemize}
\item 
The dimension $m$ of the latent noise vector $\boldsymbol{\eta} \sim N(\mathbf{0},\mathbf{I}_m)$ critically affects the expressiveness of the model. If $m$ is too small, the generator $G_\psi(\boldsymbol{\eta},\bX)$ may lack sufficient stochastic degrees of freedom to represent the full variability in $Y$ given $\bX=\bx$. If $m$ is too large, training may become more difficult, and the generator may suffer from greater stochastic variability.
\cite{zhou2023deep} does not provide a general rule for selecting the dimension $m$; instead, it noted that {\it ``The value of $m$ should be chosen on a case-by-case basis in practice.”}
\item 
Although GCDS offers much faster computation in inference (sampling) compared to the other simulation-based DDPM method reviewed in this article, this comes at the cost of reduced flexibility in capturing complex conditional structures. In addition, its training time is considerably slower than that of DDPM.
\end{itemize}

\subsection{Conditional DDPM: a diffusion-based generative approach}

The second generative simulation–based approach we consider is the conditional denoising diffusion probabilistic model. Although both DDPM and the Noise Outsourcing Lemma rely on the idea of generating randomness externally and transforming it into samples of a target conditional distribution, DDPM operates in a fundamentally different manner from GCDS. Rather than learning a one-step transformation  $\bY=G_\psi(\boldsymbol{\eta}, \bX)$, DDPM constructs a multi-step stochastic diffusion process that gradually converts data into noise and then learns a reverse denoising process capable of synthesizing new samples. Recent works, including~\cite{fu2024unveil} and~\cite{yang2025conditional}, study conditional sampling via diffusion models by defining a forward diffusion process and learning the corresponding reverse-time denoising process via score matching under a continuous-time diffusion model. For ease of exposition, and to avoid the technical overhead of continuous-time stochastic differential equations, we present a discrete-time formulation that aligns with this approach and suffices for our purposes.
The DDPM framework consists of two components:
\begin{itemize}
    \item 
    a forward diffusion process that gradually transforms the training data into random noise, and
    \item a reverse denoising process that generates data samples by progressively transforming noise back toward the data distribution.
\end{itemize} 
In the forward process, given a training sample $\bY_i$, Gaussian noise is progressively injected so that, after many steps, the distribution of the corrupted sample becomes close to a standard normal distribution.
For notational simplicity, we drop the sample index $i$, with the understanding that the following procedure is applied to each training sample.
Let $T\in\bbN$ be the number of diffusion steps, and let $\{\alpha_t\in (0,1)\}_{t=1}^T$ be a sequence of pre-specified noise schedule. Set
$\beta_t = 1-\alpha_t$, and draw i.i.d. noise vectors $\bepsilon_t \sim N(\mathbf{0},\mathbf{I}_q)$.
The forward diffusion process, which takes the form of an AR(1) model, is given by
\begin{align} \label{e:forward}
\bY_t & = \sqrt{\alpha_t}\, \bY_{t-1} + \sqrt{\beta_t}\, \bepsilon_t, \quad t = 1, \ldots, T.
\end{align}
Recursively applying \eqref{e:forward}, it leads to the following closed-form representation of the forward process:
\begin{align*}
\bY_t & = \sqrt{\alpha_t} \left(\sqrt{\alpha_{t-1}}\, \bY_{t-2} + \sqrt{\beta_{t-1}}\bepsilon_{t-1}\right) + \sqrt{\beta_t} \bepsilon_t  \\
& = \sqrt{\alpha_t\alpha_{t-1}}\, \bY_{t-2} + \sqrt{\alpha_t\beta_{t-1}}\, \bepsilon_{t-1} + \sqrt{\beta_t}\, \bepsilon_t \\
& \hspace{.2cm} \vdots \hspace{3cm} \vdots \hspace{3cm} \vdots \\
& = \sqrt{\alpha_t\cdots\alpha_1} \, \bY_0 + \sum_{j=1}^t \sqrt{\alpha_t\cdots \alpha_{j+1}\beta_j}\, \bepsilon_j
\end{align*}
with the convention $\alpha_t \alpha_{t+1}=1$ when $j=t$. Let $\wbar \alpha_t:=\prod_{s=1}^t\alpha_s$. Then the variance contributed by the noise terms is
$\sum_{j=1}^t \alpha_t\cdots \alpha_{j+1}\beta_j
= 1- \alpha_t\cdots \alpha_1 = 1-\wbar \alpha_t$.
Thus the forward process admits the compact DDPM form:
\begin{align} \label{e:art0}
\bY_t = \sqrt{\wbar\alpha_t}\, \bY_0 + \sqrt{1-\wbar\alpha_t} \, \wbar\bepsilon_t, 
\end{align}
where 
\[\wbar\bepsilon_t = (1-\wbar\alpha_t)^{-1/2}\sum_{j=1}^t \sqrt{\alpha_t\cdots \alpha_{j+1}\beta_j}\, \bepsilon_j \sim N(\mathbf{0},\mathbf{I}_q).\]
Here $\bepsilon_t$ denotes the Gaussian noise injected in the forward AR(1) process, whereas $\wbar\bepsilon_t$ is the aggregated noise appearing in the closed-form expression~\eqref{e:art0}. In contrast, $\hat\bepsilon_t$ in~\eqref{e:hatepsilon} denotes the neural network’s estimate of this aggregated noise.
Although each $\wbar\bepsilon_t$ is standard Gaussian, the sequence $\{\wbar\bepsilon_t\}$ is not independent.\ Owing to~(\ref{e:art0}),  the conditional distribution of $\bY_t$ given $\bY_0=\by_0$ converges weakly to $N(\mathbf{0}, \mathbf{I}_q)$ as $t\to\infty$. Up to this point, the conditioning variable $\bX$ has not yet appeared in the formulation. It appears naturally in the reverse denoising step through the noise-prediction model.

While the forward process diffuses the data toward a standard normal distribution, the reverse process seeks to invert this diffusion.
The reverse (denosing) process requires estimating the noise component $\wbar\bepsilon_t$ given $\bY_t=\by_t$ and $\bX=\bx$. This estimation step can be viewed as a nonparametric regression task, in which a neural network noise model $\tau_\theta$ is trained to predict the noise $\wbar\bepsilon_{i,t}$ from $(\by_{i,t}, \bx_i, t)$, with $\theta$ denoting the network parameters and $i$ indexing the training samples. Specifically, we solve
\begin{align*}  
\hat\theta := \argmin_{\{\theta: \tau_\theta\in\CF\}} \sum_{i=1}^n \sum_{t=1}^T \left\|\wbar\bepsilon_{i,t} - \tau_\theta(\by_{i,t}, \bx_i, t)\right\|^2,
\end{align*}
where $\CF$ is a suitable neural network function class. With the resulting estimate $\hat\theta$, we denote the trained noise estimation model by $\tau_{\hat{\theta}}$. For notational convenience, we write the noise estimate for the input $(\boldsymbol{y}_{t}, \boldsymbol{x},t)$ as
\begin{align} \label{e:hatepsilon} 
\hat\bepsilon_{t} := \tau_{\hat{\theta}}(\boldsymbol{y}_{t}, \boldsymbol{x},t).   
\end{align}

To describe the reverse process, we introduce the backward transition.
For simplicity, we again suppress the sample index $i$. Under the diffusion process, we have
\begin{align*}
p(\by_{t-1}|\by_t,\by_0) &= {p(\by_t|\by_{t-1}, \by_0)p(\by_{t-1}|\by_0)\over p(\by_t|\by_0)} \\
&= {p(\by_t|\by_{t-1})p(\by_{t-1}|\by_0)\over p(\by_t|\by_0)},
\end{align*}
where the second equality follows from the Markovian property of the forward diffusion process.
Combining \eqref{e:forward} and \eqref{e:art0}, together with some straightforward calculation, yields: 
\begin{align*} \label{e:xtxt-1}
\bY_{t-1}|\by_t,\by_0
\sim N\left(\mu(\by_t, \by_0), \sigma^2(t) \bf{I}_q\right),
\end{align*}
where
$$
\mu(\by_t, \by_0) = \frac{\sqrt{\alpha_t}(1-\wbar\alpha_{t-1})\by_t + \sqrt{\wbar\alpha_{t-1}}(1-\alpha_{t})\by_0}{1-\wbar\alpha_t}
$$
and
\begin{equation*} \label{posterior_var}
\sigma^2(t) = \frac{(1-\alpha_t)(1-\wbar\alpha_{t-1})}{1-\wbar\alpha_t}.
\end{equation*}
To remove dependence on the unknown $\by_0$, we substitute it with the noise expression from~\eqref{e:art0}:
\begin{equation}\label{y0_noise}
\by_0  = \frac{\by_t -\sqrt{1-\wbar\alpha_t} \, \wbar\bepsilon_t}{\sqrt{\wbar\alpha_t}},
\end{equation}
and we can estimate this noise term using the trained model \eqref{e:hatepsilon}. Substituting the expression \eqref{y0_noise} into the backward mean yields the noise-parameterized posterior mean
$$
\mu(\by_t, \by_0) = {1\over\sqrt{\alpha_t}}\left(\by_t-{1-\alpha_t\over\sqrt{1-\wbar\alpha_t}}\wbar\bepsilon_t\right).
$$
Replacing $\wbar\bepsilon_t$ by its estimator $\hat\bepsilon_t$ from \eqref{e:hatepsilon} gives the sampling rule for the reverse transition.
Thus, a sample $\by_{t-1}$ given $\by_t$ can be generated according to
\begin{align*} 
\by_{t-1} = {1\over\sqrt{\alpha_t}}\left(\by_t-{1-\alpha_t\over\sqrt{1-\wbar\alpha_t}}\hat\bepsilon_t\right) + \sigma(t) \bz,
\end{align*}
where $\bz\sim N(\mathbf{0},\mathbf{I}_q)$.

Among the methods reviewed, we provide pseudocode only for the DDPM approach, as it is the most flexible and broadly applicable across a wide range of settings. We summarize the above discussion by presenting the method in two stages: Algorithm 1 (Training) and Algorithm 2 (Sampling), which describe the procedures for learning the noise-estimation model and generating conditional samples, respectively. For computational efficiency, training does not simulate the full forward diffusion sequence over all timesteps. Instead, for each training sample indexed by $i$, a single random timestep $t_i$ is selected, and a Gaussian noise variable, denoted by $\boldsymbol{\varepsilon}_i$, is drawn to construct the noisy input $\by_{i,t_i}$. This $\boldsymbol{\varepsilon}_i$ serves solely as the synthetic noise used in the single-step training scheme and is not part of the forward AR(1) diffusion sequence discussed earlier.

Advantages:
\begin{itemize}
\item
DDPM achieves better generative quality and diversity compared with GCDS, owing to its probabilistic formulation and fine-grained denoising steps that progressively refine samples.
\item
Training is relatively stable compared with GCDS, since the DDPM objective reduces to a simple mean-squared error between the true and predicted noise. In contrast, GCDS relies on distribution matching through a min–max adversarial optimization, which is more susceptible to instability. As a result, DDPM typically exhibits more stable training and less stochastic behavior at inference than GAN-style methods such as GCDS.
\end{itemize}

Disadvantages:
\begin{itemize}
\item 
Sampling (inference) is computationally expensive because the reverse diffusion process requires many iterative denoising steps. Consequently, generation is substantially slower than GCDS in both computation time and memory usage. Faster approximate sampling methods (e.g., \citealp{song2020denoising}) are available, but they may involve trade-offs in accuracy or stability.
\end{itemize}
\begin{algorithm}
	\caption{Conditional DDPM Training}
	\footnotesize
	\begin{algorithmic}[1]\raggedright
		\State Input: dataset $\{(\bx_i,\by_{i,0})\}_{i=1}^n$, number of diffusion steps $T$
		\State Output: trained noise estimation model $\tau_{\hat\theta}(\by_t,\bx,t)$
		\State Initialize network parameters $\theta$ of $\tau_{\theta}(\by_t,\bx,t)$
		\While{not converged}
        \State Partition index set $\{1,\ldots,n\}$ into mini-batches $B_1, \ldots, B_m$
		\For{each mini-batch $B$}
		\For{each $i \in B$}
		\State Draw $t_i \sim \mathrm{Uniform}\{1,\dots,T\}$ and $\boldsymbol{\varepsilon}_i \sim N(\mathbf{0},\mathbf{I}_q)$
		\State Compute $\by_{i,t_i} = \sqrt{\wbar\alpha_{t_i}}\,\by_{i,0} + \sqrt{1-\wbar\alpha_{t_i}}\,\boldsymbol{\varepsilon}_i$
		\State Predict $\hat{\boldsymbol{\varepsilon}}_i = \tau_{\theta}(\by_{i,t_i}, \bx_i, t_i)$
		\EndFor
		\State Compute the loss $L = \frac{1}{|B|}\sum\limits_{i\in B}\left\|\boldsymbol{\varepsilon}_i - \hat{\boldsymbol{\varepsilon}}_i\right\|^2$
		\State Update $\theta \leftarrow \theta - \gamma\,\nabla_{\theta} L$
		\EndFor 
		\EndWhile
		\State \textbf{Return:} $\tau_{\hat\theta}$
\end{algorithmic}
\end{algorithm}

\begin{algorithm}
\caption{Conditional DDPM Sampling}
\footnotesize
\begin{algorithmic}[1]\raggedright
\State{Input: predictor $\bx$ and trained noise model $\tau_{\hat\theta}$}
\State{Output: conditional sample $\by_0$}
\State{Draw $\by_T \sim N(\mathbf{0},\mathbf{I}_q)$}
\For{$t=T,\ldots,1$}
\State Compute predicted noise $\hat\bepsilon_t = \tau_{\hat\theta}(\by_t,\bx,t)$
\State Draw $\bz \sim N(\mathbf{0},\mathbf{I}_q)$ if $t>1$, else $\bz={\bf 0}$
\State Update by $\by_{t-1}={1\over\sqrt{\alpha_t}}\left(\by_t-{1-\alpha_t\over\sqrt{1-\wbar\alpha_t}}\hat\bepsilon_t\right) + \sigma(t) \bz$
\EndFor
\State{return $\by_0$}
\end{algorithmic}
\end{algorithm}


\section{Numerical studies}\label{s:simulations}

This section presents a systematic comparison of the approaches introduced in Section~\ref{s:approaches} through controlled simulation experiments designed to evaluate their empirical performance under various settings. 

\subsection{Simulation models}
In order to conduct a broad and informative comparison among approaches grounded in distinct methodologies, we consider multiple simulation scenarios that exhibit a wide range of conditional behaviors. These models cover a wide spectrum of conditional behaviors, ranging from smooth, homoscedastic, and Gaussian settings to scenarios exhibiting heteroscedasticity, high skewness, heavy tails, covariate-dependent mixtures, and latent mixture structures. By these different features of simulation models, we illuminate the types of distributional characteristics that each method is able to capture, as well as those that present greater challenges, thereby providing a more balanced assessment of their respective strengths and limitations.

We consider the following simulation models.
\begin{enumerate}[label=M\arabic*, ref=M\arabic*]
\item \label{M1:sinx} 
This is Example~2 of \cite{hall2005approximating}. Let
\begin{align}\label{e:M1} 
    Y = 0.5 \sum_{j=1}^4 \sin X_j + \varepsilon,
\end{align}
where $X_j, j=1,\ldots,4,$ and $\varepsilon$ are i.i.d.\ $N(0,1)$. This model features a smooth conditional mean and homoscedastic Gaussian noise. 

\item \label{M2:sinx_redundant} 
This model is identical to \ref{M1:sinx} except that the number of predictors is increased to $10$; that is, 
$\bX = (X_1,\ldots,X_{10})^\top \in \mathbb{R}^{10}$ with $X_j \stackrel{\mathrm{i.i.d.}}{\sim} N(0,1)$ and independent $\varepsilon \sim N(0,1)$. 
The response $Y$ is still generated by~\eqref{e:M1}, so $X_5,\ldots,X_{10}$ are redundant predictors. This setting examines whether the methods remain stable in the presence of irrelevant predictors.

\item \label{M3:sinx_correlated} 
This model is based on \ref{M1:sinx}, except that $\bX = (X_1,\ldots,X_4)^\top$ follows a multivariate normal 
distribution $N(\mathbf{0},\mathbf{\Sigma})$ with $\mathbf{\Sigma}_{ij} = 0.5^{|i-j|}$ for $i,j=1,\ldots,4$, independent of the error term $\varepsilon \sim N(0,1)$.
This model introduces correlation among predictors.

\item \label{M4:sinx_latent_discrete} 
This model extends \ref{M1:sinx} by introducing latent sign variables. 
Let $\bZ=(Z_1,\dots,Z_4)^\top$, where $Z_j$ are i.i.d. with $\Pr(Z_j=1)=\Pr(Z_j=-1)=1/2$, independent of $\bX$ and $\varepsilon$. Define
\begin{align}\label{e:M1_withZ}
    Y = 0.5 \sum_{j=1}^4 \sin (Z_jX_j) +\varepsilon.
\end{align}
$\bZ$ serves as an unobserved latent factor that randomly flips the sign of each $X_j$. The marginal distribution of $Y$ in \eqref{e:M1_withZ} remains the same as that of $Y$ in \eqref{e:M1}; however, their conditional distributions of $Y\mid \bX$ differ.
Since $\bZ$ is latent and unavailable to the estimation procedures, this model provides a useful setting for assessing the robustness of different methods to unobserved heterogeneity.

\item \label{M5:sinx_latent_continuous} 
This model extends \ref{M1:sinx} by introducing predictor-specific latent scale variables. The response is generated by the same formula as in \eqref{e:M1_withZ}, but with $Z_j \stackrel{\mathrm{i.i.d.}}{\sim} \mathrm{Uniform}(0,1)$, independent of 
$\bX$ and $\varepsilon$.
The latent variables $Z_j$ introduce random continuous modulation of the 
predictor effects, creating feature-wise nonlinear distortions.  
As a result, the conditional distribution of $Y \mid \bX$ becomes more 
heterogeneous and distinctly non-Gaussian, exhibiting varying local curvature 
and scale depending on the latent realization of $\bZ$.

\item \label{M6:heteroscedastic_noise} 
This model corresponds to Example~1 of \cite{yang2025conditional} and Example~2 of \cite{zhou2023deep}. Let
\begin{align*}
Y &= X_1^2 + \exp(X_2 + X_3/3) + X_4 - X_5 \\
& \hspace{.5cm} + (0.5+ X_2^2/2+ X_5^2/2) \times \varepsilon, 
\end{align*}
where $\varepsilon \sim N(0,1)$.
This model exhibits heteroscedasticity arising from nonlinear covariate effects on the noise scale.

\item \label{M7:heteroscedastic_covariate-dependent_mixture_noise} 
This model is a milder counterpart to \ref{M8:heteroscedastic_latent_mixture_noise} described below. It retains the same multiplicative structure but uses a milder and covariate-dependent noise, thereby yielding an easier problem. For $\bX\in\mathbb{R}^{30}$, let
\[
m(\bX) = 5+ X_1^2/3+ X_2^2 + X_3^2 + X_4 + X_5,
\]
and define
\[
Y = m(\bX)\times \exp(0.25\,\varepsilon), 
\]
where 
$\varepsilon \mid X_1 \sim \pi(X_1)\times N(-1,1) + \{1-\pi(X_1)\}\times N(1,1)$,
and $\pi(x) = \mathrm{logit}^{-1}(\kappa x)$. The parameter $\kappa>0$ governs the sensitivity of the mixing weights to $X_1$: larger values produce sharper changes in the mixing weight $\pi(X_1)$ and hence stronger deviations from the pure scaling structure in \ref{M8:heteroscedastic_latent_mixture_noise}, whereas smaller values yield milder shape differences and a problem closer to the scaling model. In other words, $\kappa$ tunes how non-scaling and how heterogeneous the conditional density is. 
Here we take $\kappa=0.5$.
Compared with \ref{M8:heteroscedastic_latent_mixture_noise}, the noise variation is milder.

\item \label{M8:heteroscedastic_latent_mixture_noise} 
This model follows Example~2 of \cite{yang2025conditional} and Example~3 of \cite{zhou2023deep}. Using the same function $m(\bX)$ as in \ref{M7:heteroscedastic_covariate-dependent_mixture_noise}, define
\[
Y = m(\bX) \times \exp(0.5\,\varepsilon),
\]
where $\varepsilon \sim I(U<0.5)\times N(-2,1) + I(U>0.5)\times N(2,1)$ with $U\sim 
\text{Uniform}(0,1)$ and $\bX\in\bbR^{30}$.
The latent mixture inside the exponential induces stronger heteroscedasticity and shape variation than in \ref{M7:heteroscedastic_covariate-dependent_mixture_noise}.

\item \label{M9:skewed_heavy-tailed_bimodal} 
Let $X_1$ and $\varepsilon$ be i.i.d.\ $N(0,1)$, and let $c>0$
control the oscillation frequency of $\sin (c X_1 + \varepsilon)$ in $X_1$. Define
\[
Y = g\left(\sin (c X_1 + \varepsilon)\right)~
\mbox{with $g(u)=\exp{(u)}$},
\]
and set $c=20$.
The composition of a highly oscillating sinusoid with the exponential mapping produces highly skewed and heavy-tailed conditional distributions.

\item \label{M10:multivariate} 
This model evaluates multivariate conditional distribution estimation for $\bY\in\mathbb{R}^7$. Let
\begin{align*}
\bY & = \Big(X_1^2,X_2^2,X_3^2,X_4^2, 
X_5^2, \\
& \hspace{1cm} \exp(X_2+X_5/3),\sin(X_4+X_5)\Big)^\top + \boldsymbol{\varepsilon},
\end{align*}
where $\bX\sim N(\mathbf{0},\mathbf{I}_5)$ and $\boldsymbol{\varepsilon}\sim N(\mathbf{0},\mathbf{I}_7)$.
The heterogeneous nonlinear components in $\bY$ create a nontrivial multivariate conditional structure useful for comparing GCDS and DDPM.
\end{enumerate}

\begin{figure}[htbp]
  \centering
  \includegraphics[width=0.95\textwidth]{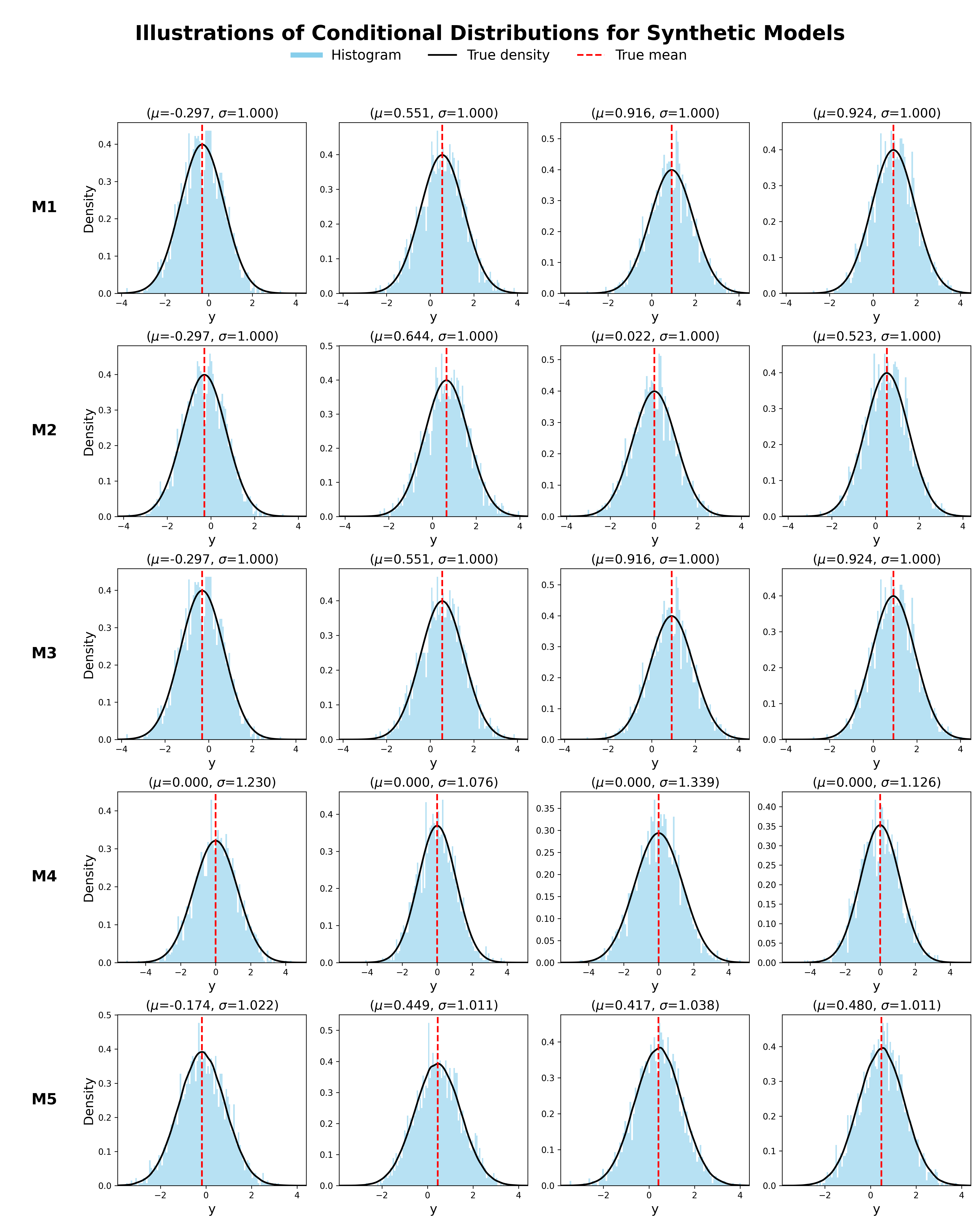}
  \caption{True conditional densities for models M1-M5.}
  \label{fig:cond-m1-4}
\end{figure}

\begin{figure}[htbp]
  \centering
  \includegraphics[width=0.95\textwidth]{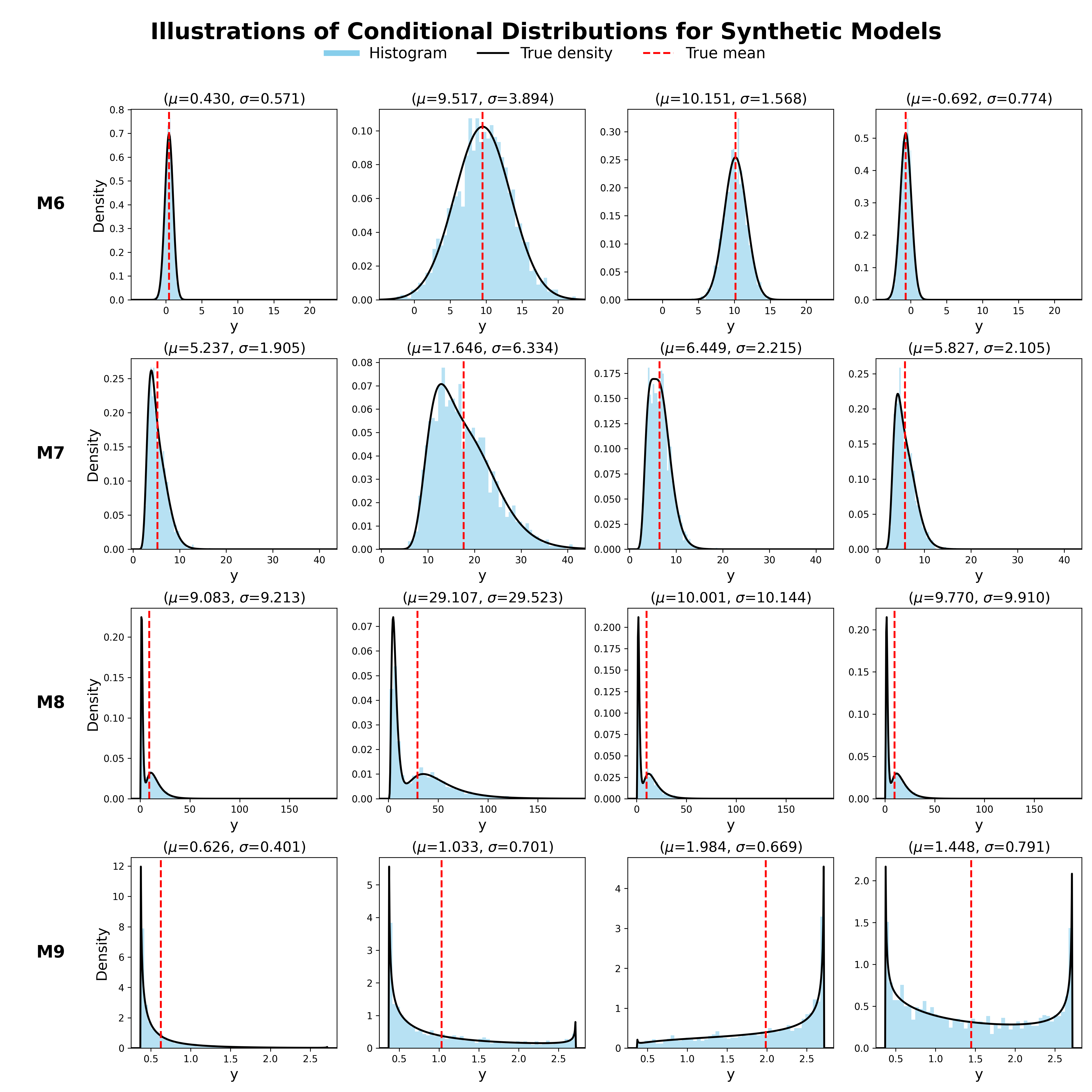}
  \caption{True conditional densities for models M6-M9.}
  \label{fig:cond-m4a-8}
\end{figure}

To illustrate the behavior of the synthetic models (excluding \ref{M10:multivariate}, which is multivariate), we display their true conditional densities in Figures~\ref{fig:cond-m1-4}-\ref{fig:cond-m4a-8}.
For each model, four predictor values $\bx$ are drawn at random, and the true conditional density is evaluated directly from the data-generating mechanism.
Among the proposed synthetic models, \ref{M1:sinx}–\ref{M4:sinx_latent_discrete} and \ref{M6:heteroscedastic_noise}–\ref{M8:heteroscedastic_latent_mixture_noise} admit closed-form conditional densities.
In contrast, the conditional densities of \ref{M5:sinx_latent_continuous} and \ref{M9:skewed_heavy-tailed_bimodal} are unavailable for closed form and are therefore evaluated approximately.
Each panel shows the true conditional density (black curve), together with the true conditional mean marked by a red dashed line.

\subsection{Simulation design}

We conducted 10 independent simulation runs for each algorithm to assess performance variability. The number of runs was limited to 10 due to the high computational cost and the large number of methods being compared. Nevertheless, the standard deviations reported in Table~\ref{tab:method_comparison} indicate modest variability across runs, suggesting that this choice is reasonable.
In each run, the model was trained on $5{,}000$ conditional sample pairs, validated on $2{,}000$ samples, and evaluated on $2{,}000$ previously unseen test samples.  
For each test input, we generated $2{,}000$ samples from the estimated conditional distribution and computed the evaluation metrics described in Section~\ref{s:metrics}.   
Note that while it is possible to compute the metrics numerically for both the Hall and Yao and FlexCode methods, we chose to standardize the process by conducting simulations for all methods.

The normalization schemes used in the experiments are described below.
For FlexCode and DeepCDE, we follow the normalization settings in the original authors’ implementations.
For GCDS and DDPM, which are implemented in this work, standard normalization is applied to facilitate stable training.
All generated samples are rescaled to their original physical scale prior to computing the evaluation metrics.

\begin{itemize}
\item
FlexCode: the response $Y$ is scaled to $[0,1]$.
\item 
DeepCDE: the response $Y$ is scaled to $[0,1]$, and the predictors $\bX$ are standardized coordinate-wise to have mean zero and unit variance.
\item 
GCDS: the response $Y$ is standardized to have mean zero and unit variance.
\item 
DDPM: the response $Y$ is standardized to have mean zero and unit variance.
\end{itemize}

The configurations for each method are summarized as follows.
\begin{itemize}
    \item 
    For Hall and Yao, we set the radius to $r^\ast=1$. For bandwidth selection in each model, we carried out a grid search to choose $(h, H)$, corresponding to the bandwidths for estimating $\bv$ and $F\left(y \mid \hat{\bv}^\top \bx\right)$, over a coarse grid, where $h \in \{0.1, 0.3, \ldots, 1.1\}$ and $H \in \{h+ 0.1, h+0.3, h+0.5, \ldots, 1.2\}$. The grid search was conducted using a single training sample of size $5{,}000$, and the selected pair was chosen by minimizing the Wasserstein distance on a validation sample of size $2{,}000$. After obtaining the optimal pair $(\hat{h}, \hat{ H})$, we fixed it and applied the Hall and Yao method to independent training and test samples for evaluation. Although the selected pair $(\hat{h}, \hat{ H})$ may vary across different data realizations, the resulting inferential performance averaged over 10 independent runs remains very similar.
    \item For FlexCode, we employed a cosine basis together with random forest regression. The contributing basis functions were selected, based on their importance in minimizing the empirical conditional density estimation loss, from a maximum of 31 candidate basis functions. 
    \item For DeepCDE, we used a set of 31 cosine basis functions and a learning rate of $10^{-4}$ for all models, with early stopping based on the validation loss and a patience of 20 epochs.
    \item For GCDS, each model was trained for 500 epochs, using a learning rate of $3 \times 10^{-4}$ for both the generator and discriminator in \ref{M10:multivariate}, and $10^{-4}$ for both networks in all remaining models.
    \item For DDPM, the diffusion process employed a linear noise schedule. The models were trained for $50$ epochs with an initial learning rate of $10^{-2}$ and a drop factor of $0.5$ every $10$ epochs.
\end{itemize}

\subsection{Model implementation details}

We next describe the implementation details of the learning-based methods, focusing on the neural network architectures and the implementation environment.

\subsubsection{Neural network architectures}
{\it DeepCDE network architecture.}
For DeepCDE, we employ a three-hidden-layer MLP (with width $32$-$64$-$32$ and GELU activations) to extract nonlinear representations from the predictors. A final linear layer (the CDE layer) maps the $32$-dimensional representation to $J$ basis coefficients $\{\beta_j(\bx)\}_{j=1}^{J}$ corresponding to the chosen basis system. The entire network is trained by minimizing the CDE loss. This compact multilayer perceptron enables DeepCDE to capture nonlinear dependencies between predictors and efficiently estimate the conditional density function
$\hat{f}_{Y\mid \bX}(y \mid \bx) = \sum_{j=1}^{J} \hat{\beta}_j(\bx)\,\varphi_j(y)$,
where $\{\varphi_j\}_{j=1}^{J}$ denotes the chosen orthonormal basis in $L^2(\mathbb{R})$. The number of basis functions $J$ is a tunable hyperparameter. In the simulation studies, we set $J=31$. In addition, we use orthonormal cosine basis for $\varphi$.

{\it GCDS network architecture.}
The generator takes as input the concatenation of the predictors and a latent noise vector.
This input is passed through a single hidden-layer MLP with width $50$ and a ReLU activation to generate a conditional sample of dimension $q$.
The discriminator receives the concatenation of the predictors and a response, and processes it through a two-hidden-layer MLP with widths $50$ and $25$, each followed by ReLU activations. A final linear layer maps the 25-dimensional representation to a single scalar output, which serves as the discriminator’s score indicating how likely the predictor–response pair is to have come from the true conditional distribution.
  
{\it DDPM network architecture.}
We use a conditional MLP that takes as input the concatenation of the predictors, a noised version of the response, and the scalar diffusion time. 
The network consists of two hidden layers with widths $50$ and $25$, each followed by a ReLU activation, and ends with a linear output layer that maps to a vector in $\bbR^q$.
The network parameters are estimated by minimizing the mean squared error between the true noise and the predicted noise. 

\subsubsection{Implementation environment}
All learning-based methods were implemented in a common software environment using standard deep learning libraries.
To ensure fair comparisons, all experiments were conducted on the same computing platform and executed exclusively on the CPU, with no GPU acceleration.
Specifically, experiments were run on a server equipped with an AMD EPYC 7742 processor (64 cores, 128 threads) and 1~TiB of RAM.

\subsection{Evaluation metrics}\label{s:metrics}
The first issue to address is the choice of discrepancy measures between the estimated and the true conditional distributions. 
There are many possibilities depending on what aspect the conditional distribution is being evaluated. For example, 
\cite{zhou2023deep} focus on how well the true conditional mean and conditional variance can be estimated by the simulated conditional samples, using the following mean squared errors:
\begin{align*}
\begin{split}
&\text{MSE (mean)} = {1\over k} \sum_{i=1}^k \left(\widehat{\E}(Y\mid \bx_i)-\E(Y\mid \bx_i)\right)^2, \\
&\text{MSE (sd)} = {1\over k} \sum_{i=1}^k \left(\widehat{\text{SD}}(Y\mid \bx_i)-\text{SD}(Y\mid \bx_i)\right)^2,
\end{split}
\end{align*}
where $\{\bx_1,\ldots, \bx_k\}$ is a test set, and, $\widehat{\E}(Y\mid \bx_i)$, $\widehat{\text{SD}}(Y\mid \bx_i)$, $i = 1,\ldots,k$, are computed from simulated samples of $Y$ at each $\bx_i$. 
While the mean and standard deviation fully determine a normal distribution, they are not ideal measures for judging the discrepancy between two distributions in general. 
A broader discussion of possible distance measures between distributions is provided in \cite{ramdas2017wasserstein}.

In addition to the MSE of the conditional mean and standard deviation, we also adopt the Wasserstein distance as our primary discrepancy measure.
For $r \in[1, \infty)$ and Borel probability measures $\mu_1, \mu_2$ on $\mathbb{R}^q$ with finite $r$-moments, their $r$-Wasserstein distance is defined as
$$
W_r(\mu_1, \mu_2)=\left(\inf _{\pi \in \Gamma(\mu_1, \mu_2)} \int_{\mathbb{R}^q \times \mathbb{R}^q}\|\by-\by'\|^r d \pi(\by,\by')\right)^{1 / r},
$$
where $\Gamma(\mu_1, \mu_2)$ is the set of all joint probability measures $\pi$ on $\mathbb{R}^q \times \mathbb{R}^q$ whose marginals are $\mu_1$ and $\mu_2$. That is, for all subsets $A^\ast \subset \mathbb{R}^q$, we have $\pi\left(A^\ast \times \mathbb{R}^q\right)=\mu_1(A^\ast)$ and $\pi\left(\mathbb{R}^q \times A^\ast\right)=\mu_2(A^\ast)$. If $q=1$, then the $r$-Wasserstein distance simplifies to
$$
W_r(\mu_1, \mu_2)= \left(\int_0^1 |F_1^{-1}(u)-F_2^{-1}(u)|^r du\right)^{1/r},
$$
where $F_1^{-1}$ and $F_2^{-1}$ are the quantile functions corresponding to $\mu_1$ and $\mu_2$, respectively. In practice, the Wasserstein distance $W_r(\mu_1, \mu_2)$ can be approximated by $W_r(\mu_{1,n}, \mu_{2,n})$, where $\mu_{1,n}$ and $\mu_{2,n}$ are empirical measures based on finite samples. The \texttt{scipy.stats} library provides the function \texttt{wasserstein\_\allowbreak distance} for computing the 1-Wasserstein distance for univariate distributions, as well as \texttt{wasserstein\_\allowbreak distance\_\allowbreak nd} for multivariate distributions. However, as noted by \cite{ramdas2017wasserstein}, the empirical estimate $W_r(\mu_{1,n}, \mu_{2,n})$ may converge slowly to the true Wasserstein distance $W_r(\mu_{1}, \mu_{2})$ when $q>1$. To obtain a computationally efficient approximation, we instead compute the sliced Wasserstein distance between the empirical distributions using the function \texttt{sliced\_\allowbreak wasserstein\_\allowbreak distance} from the \texttt{POT} (Python Optimal Transport) library \citep{flamary2021pot}. The sliced Wasserstein distance approximates the multivariate Wasserstein distance by averaging 1D Wasserstein distances of the distributions projected onto multiple random directions~\citep{bonneel2015sliced}, which is fast to compute and effective in practice.

\subsection{Results}
To assess the comparative performance of conditional distribution estimation, 
we evaluate the three metrics described in Section~\ref{s:metrics} for each method across a series of simulated examples. Table~\ref{tab:method_comparison} summarizes the results based on 10 simulation runs. Boldface indicates the method that achieves the best overall performance in terms of distributional accuracy, as reflected primarily by the Wasserstein distance together with the conditional mean MSE. In some settings, a method may exhibit smaller variability (e.g., lower MSE of the conditional standard deviation) while remaining farther from the target conditional distribution.

\ref{M1:sinx} serves as a simple baseline model with smooth nonlinear main effects and homoscedastic Gaussian noise. \ref{M2:sinx_redundant}-\ref{M5:sinx_latent_continuous} extend \ref{M1:sinx} by introducing additional structural complexity: \ref{M2:sinx_redundant} adds redundant predictors, \ref{M3:sinx_correlated} introduces correlated predictors, \ref{M4:sinx_latent_discrete} incorporates latent sign variables that give rise to multiple regression-function branches, and \ref{M5:sinx_latent_continuous} introduces continuous latent scales. These settings form an increasingly challenging sequence for conditional distribution estimation. 
With the exception of the Hall and Yao method on \ref{M1:sinx}–\ref{M3:sinx_correlated} and \ref{M5:sinx_latent_continuous}, all approaches perform well on \ref{M1:sinx}–\ref{M5:sinx_latent_continuous}. Notably, Hall and Yao performs particularly well on \ref{M4:sinx_latent_discrete}. Among these, DDPM achieves the strongest overall performance on \ref{M1:sinx}–\ref{M3:sinx_correlated}, and the Hall and Yao method and DeepCDE exhibit the strongest performance on \ref{M4:sinx_latent_discrete} and \ref{M5:sinx_latent_continuous}, respectively, according to the three evaluation metrics.

\ref{M6:heteroscedastic_noise}-\ref{M9:skewed_heavy-tailed_bimodal} represent settings with heteroscedastic noise, and \ref{M7:heteroscedastic_covariate-dependent_mixture_noise}-\ref{M9:skewed_heavy-tailed_bimodal} additionally produce non-Gaussian conditional distributions. \ref{M6:heteroscedastic_noise} allows the noise scale to vary with the predictors. In \ref{M7:heteroscedastic_covariate-dependent_mixture_noise}, the conditional distribution is a non-Gaussian mixture induced by a covariate-dependent Gaussian mixture in the noise term. \ref{M8:heteroscedastic_latent_mixture_noise} employs a latent-weight Gaussian mixture, which produces more complex conditional densities, often exhibiting multimodality and pronounced shape variation. In \ref{M9:skewed_heavy-tailed_bimodal}, the nonlinear transformation leads to highly skewed, heavy-tailed, or bimodal conditional distributions. Relative to \ref{M8:heteroscedastic_latent_mixture_noise} and \ref{M9:skewed_heavy-tailed_bimodal}, the settings in \ref{M6:heteroscedastic_noise} and \ref{M7:heteroscedastic_covariate-dependent_mixture_noise} are comparatively simpler for conditional distribution estimation. DDPM and FlexCode yield the strongest performance on \ref{M6:heteroscedastic_noise}-\ref{M7:heteroscedastic_covariate-dependent_mixture_noise} and \ref{M8:heteroscedastic_latent_mixture_noise}-\ref{M9:skewed_heavy-tailed_bimodal}, respectively. 

\ref{M10:multivariate} is a multivariate example designed to compare GCDS and DDPM; see the next two paragraphs for the reasons why the other methods are not
considered. In this setting, DDPM performs better than GCDS. More broadly, across all simulation settings except~\ref{M9:skewed_heavy-tailed_bimodal}, DDPM generally outperforms GCDS under the three evaluation metrics considered for conditional distribution estimation. In~\ref{M9:skewed_heavy-tailed_bimodal}, the two methods exhibit comparable performance, with GCDS performing slightly better.
A key reason is that DDPM employs a multi-step diffusion-based generative mechanism, which yields more stable and effective learning. By contrast, GCDS involves a min-max adversarial optimization, which is known to be more sensitive to training instability (see, e.g., \citealp{goodfellow2014generative,dhariwal2021diffusion}; empirical evidence in the conditional settings was also reported in \citealp{han2022card}).

The Hall and Yao method is a dimension-reduction approach and, as a result, offers less flexibility than the other methods for conditional distribution estimation. 
Note that the results based on the Hall and Yao method are not reported for models \ref{M7:heteroscedastic_covariate-dependent_mixture_noise}, \ref{M8:heteroscedastic_latent_mixture_noise}, and \ref{M10:multivariate}. This is because it was specifically developed for the scalar-response setting, and the computational burden grows rapidly with the predictor dimension $p$, making it practically infeasible for higher-dimensional settings due to memory limitations. In our simulation study, because no publicly available implementation of this method exists, we developed our own based on the original paper. Following their formulation, the algorithm requires constructing a full $p$-dimensional Cartesian grid of centers $\{A_{\boldsymbol{\delta}}\}$ within the hypercube $[-r^\ast,\,r^\ast]^p$ with grid spacing~$h$. The exponential growth of the grid size renders the full grid computationally infeasible even for moderate dimensions. For instance, in \ref{M7:heteroscedastic_covariate-dependent_mixture_noise} 
and \ref{M8:heteroscedastic_latent_mixture_noise} with dimension $p=30$, this would require on the order of $8\times10^{14}$~bytes of memory. 
To alleviate this issue, we experimented with a random sampling scheme that replaced the exhaustive mesh construction by drawing a fixed number of centers uniformly from $[-r^\ast,\,r^\ast]^p$. While this strategy  substantially reduced the memory requirement, empirical results indicated that it led to degraded performance, suggesting that the random sampling strategy was insufficient to capture the fine structural dependencies represented by the full grid.

Results based on FlexCode, and DeepCDE are also not reported for \ref{M10:multivariate} since the publicly available implementations of FlexCode and DeepCDE currently support only the univariate response case.
FlexCode, which estimates each coefficient $\bm{\beta}_{j}(\bx)$ by regressing $\varphi_j(Y)$ on $\bX$ using methods such as random forests or $k$-nearest neighbors, performs poorly under conditional heteroscedasticity, where the variability of $\varphi_j(Y)\mid\bX=\bx$ depends strongly on $\bx$. DeepCDE, in contrast, learns all coefficients simultaneously through the optimization of a single neural-network objective, allowing it to better adapt to heteroscedastic structure in the data. Similar behavior is observed under \ref{M6:heteroscedastic_noise} and \ref{M7:heteroscedastic_covariate-dependent_mixture_noise}. 

However, FlexCode is not constrained by a fixed neural network architecture and remains a reasonably flexible approach when a moderate number of basis functions is employed together with an appropriate regression method, even for estimating conditional densities that exhibit complex shapes (e.g. highly skewed, multimodal, or heavy-tailed). This flexibility helps to explain its stronger performance relative to the neural-network-based methods in \ref{M8:heteroscedastic_latent_mixture_noise} and \ref{M9:skewed_heavy-tailed_bimodal}. We suggest that increasing the architectural complexity of DeepCDE, GCDS and DDPM would likely improve their ability to learn the latent mixtures and more complex distributional structures present in these models.

From a computational perspective, Table~\ref{tab:epoch_time} reports average epoch times for DeepCDE, GCDS, and DDPM. 
Overall, DeepCDE achieves the fastest per-epoch training times, while GCDS is consistently the slowest due to its adversarial min-max optimization. 
DDPM lies in between: its per-epoch training cost is moderate and considerably lower than that of GCDS, but its sampling procedure is much more expensive, as reflected in the sampling times reported in Table~\ref{tab:method_comparison}. 
This gap arises because each conditional sample from DDPM requires running the full reverse diffusion chain, whereas GCDS generates samples in a single forward pass of the generator network. 
These findings are consistent with the methodological discussion in Section~\ref{s:approaches}: DDPM often attains the best or near-best statistical performance, but its generation is substantially slower than GCDS, unless one resorts to faster approximate samplers (e.g., \citealp{song2020denoising}) that may trade off some accuracy or stability.

\begin{table*}[htbp]
\centering
\caption{Performance comparison of five methods across selected data models. Values shown as mean (standard deviation) over 10 simulation runs. Lower values indicate better performance for all metrics.}
\label{tab:method_comparison}
\footnotesize
\setlength{\tabcolsep}{4pt}
\begin{tabular}{@{}llccclll@{}}
\toprule
\textbf{Model} & \textbf{Method} & \textbf{MSE Mean} & \textbf{MSE Std} & \textbf{W-1}  && \textbf{Training Time} & \textbf{Sampling Time}\\
\midrule
M1    
& Hall \& Yao   & 0.3023 (0.1289)  & 0.0257 (0.0155)  & 0.4503 (0.1187)  && 38.1 (39.1)  & 18.0 (0.6)  \\
        & FlexCode      & 0.0415 (0.0041) & 0.0262 (0.0065) & 0.2042 (0.0113) && 32.1 (0.3)  & 12.5 (0.5) \\
        & DeepCDE       & 0.0422 (0.0064) & 0.0173 (0.0027) & 0.1744 (0.0119) && 34.6 (10.8)  & 4.1 (0.6)  \\
        & GCDS          & 0.0459 (0.0094) & 0.0049 (0.0015) & 0.1667 (0.0154) && 287.2 (2.5) & 5.9 (0.3)  \\
        &{\bf DDPM } & 0.0281 (0.0149) & 0.0056 (0.0033) & 0.1386 (0.0286) && 22.2 (0.6) & 211.1 (1.5) \\        
\midrule
M2   
& Hall \& Yao   & 0.3887 (0.0545)  & 0.0338 (0.0074)  & 0.5262 (0.0410)  &&  371.7 (411.1) & 18.1 (0.6)  \\
        & FlexCode      & 0.0475 (0.0037) & 0.0298 (0.0039) & 0.2208 (0.0068) && 40.1 (4.7)  & 13.5 (1.6) \\
        & DeepCDE       & 0.0705 (0.0035) & 0.0327 (0.0092) & 0.2415 (0.0087) && 28.3 (7.6)  & 4.2 (0.4)  \\
        & GCDS          & 0.1185 (0.0136) & 0.0059 (0.0016) & 0.2686 (0.0156) && 287.2 (3.0) & 6.2 (0.4)  \\
        &{\bf DDPM } & 0.0694 (0.0127) & 0.0071 (0.0015) & 0.2139 (0.0186) && 22.0 (0)    & 216.4 (1.7) \\
\midrule
M3    
        & Hall \& Yao   & 0.4392 (0.3412) & 0.0615 (0.0557) & 0.5230 (0.2635) && 74.5 (70.5) & 27.2 (5.1) \\
        & FlexCode      & 0.0395 (0.0023) & 0.0397 (0.0082) & 0.2119 (0.0069) && 38.0 (3.0)  & 13.1 (0.7) \\
        & DeepCDE       & 0.0314 (0.0046) & 0.0259 (0.0077) & 0.1613 (0.0102) && 27.7 (4.8)  & 3.9 (0.3)    \\
        & GCDS          & 0.0530 (0.0090) & 0.0076 (0.0060) & 0.1944 (0.0129) && 285.5 (2.4) & 5.7 (0.5)  \\
        &{\bf DDPM } & 0.0256 (0.0092) & 0.0040 (0.0010) & 0.1353 (0.0174) && 22.2 (0.4) & 212.3 (1.5) \\
\midrule
M4    
& {\bf Hall \& Yao}   & 0.0028 (0.0009)   &  0.0061 (0.0007)  & 0.0826 (0.0047)   && 33.8 (30.8)   &  19.2 (0.6)  \\
        & FlexCode      & 0.0060 (0.0017) & 0.0096 (0.0031) & 0.1047 (0.0097) && 36.1 (7.0) & 16.3 (0.7) \\
        & DeepCDE       & 0.0027 (0.0013) & 0.0139 (0.0031) & 0.1003 (0.0089) && 13.3 (1.9) & 2.5 (0.5) \\
        & GCDS          & 0.0191 (0.0160) & 0.0076 (0.0014) & 0.1413 (0.0337) && 273.0 (2.5) & 4.8 (0.4) \\
        & DDPM    & 0.0105 (0.0062) & 0.0073 (0.0012) & 0.1113 (0.0183) && 23.2 (0.9) & 213.5 (1.8) \\
\midrule 
M5    
& Hall \& Yao   &  0.1176 (0.0481)    &   0.0047 (0.0020)    &   0.2793 (0.0682)   &&  47.0 (31.1)   &  19.0 (1.6)   \\
        & FlexCode      & 0.0198 (0.0026) & 0.0107 (0.0027) & 0.1458 (0.0051) && 32.1 (0.3)  & 12.6 (0.5) \\
        &{\bf DeepCDE}  & 0.0090 (0.0024) & 0.0072 (0.0028) & 0.1027 (0.0099) && 21.6 (5.2) & 4.0 (0)   \\
        & GCDS          & 0.0235 (0.0186) & 0.0033 (0.0014) & 0.1353 (0.0415) && 291.4 (3.9) & 5.3 (0.5)  \\
        & DDPM       & 0.0179 (0.0133) & 0.0057 (0.0045) & 0.1234 (0.0362) && 22.1 (0.7) & 212.0 (1.5)  \\
        
\midrule 
M6    
& Hall \& Yao   &  9.6668 (0.9152)   &  1.2568 (0.0583)   &  2.2461 (0.1026)   &&  58.4 (59.7)   &   18.7 (0.5)  \\
        & FlexCode      & 4.2328 (0.4169) & 34.3715 (6.5531) & 2.6339 (0.1467) && 34.0 (2.0)  & 13.5 (2.0) \\
        & DeepCDE       & 0.9187 (0.3077) & 4.0832 (1.2943) & 0.7417 (0.0692) && 71.0 (3.8)  & 3.0 (0)  \\
        & GCDS          & 0.5603 (0.2113) & 0.4041 (0.1037)  & 0.5153 (0.0581) && 285.2 (0.9) & 5.4 (0.5)  \\
        &{\bf DDPM } & 0.2875 (0.0994) & 0.0785 (0.0288) & 0.3007 (0.0367) && 22.0 (0.5) & 212.8 (2.1)  \\

\midrule 
M7      & FlexCode      & 2.3556 (0.1946) & 4.6753 (0.8336) & 1.5303 (0.0627) && 42.0 (0.5) & 14.0 (0) \\
        & DeepCDE       & 1.8959 (0.2948) & 3.4071 (0.6394) & 1.2471 (0.0580) && 34.6 (4.8) & 3.1 (0.3) \\
        & GCDS          & 2.3605 (0.2883) & 1.2614 (0.4157) & 1.5737 (0.0968) && 311.4 (47.5) & 4.1 (0.7) \\
        & {\bf DDPM }      & 0.9465 (0.0984) & 0.3384 (0.0483) & 0.8245 (0.0281) && 23.1 (0.7) & 216.0 (3.4) \\
\midrule 
M8      & {\bf FlexCode}      & 6.9075 (1.7249)  & 18.1483 (2.7374)   & 2.7234 (0.1746) && 42.2 (1.9)  & 13.3 (0.8) \\
        & DeepCDE       & 26.9664 (3.7139) & 68.4103 (25.1083) & 4.5726 (0.3623) && 19.1 (3.1)  & 3.0 (0)  \\
        & GCDS          & 18.0615 (2.6195) & 14.3314 (0.9252)   & 4.0671 (0.2768) && 292.2 (3.3) & 4.8 (0.6) \\
        & DDPM       & 12.3189 (1.3913) & 11.7034 (1.4097) & 3.3847 (0.1341) && 22.1 (0.3) & 211.2 (6.6)  \\

\midrule
M9   
& Hall \& Yao   &  0.2168 (0.0157)    &  0.3093 (0.1157)    &   0.5684 (0.0545)    &&  5.2 (0.4)   &   15.0 (0.7)   \\
        &{\bf FlexCode} & 0.1125 (0.0054) & 0.0383 (0.0024) & 0.2955 (0.0106) && 32 (0.5) & 15.8 (0.4) \\
        & DeepCDE       & 0.2376 (0.0025) & 0.0503 (0.0024) & 0.4424 (0.0028) && 13.9 (2.4) & 3.8 (0.4) \\
        & GCDS          & 0.2377 (0.0028) & 0.0449 (0.0063) & 0.4428 (0.0027) && 275.9 (1.5) & 5.1 (0.3) \\
        & DDPM       & 0.2393 (0.0053) & 0.0466 (0.0050) & 0.4477 (0.0046) && 22.6 (0.5) & 224.8 (61.0) \\

\midrule
M10     & GCDS          & 0.6226 (0.1277) & 0.1414 (0.0136) & 0.6171 (0.0648) && 283.8 (2.1) & 5.8 (0.4)  \\
        &{\bf DDPM } & 0.1244 (0.0469) & 0.0282 (0.0044) & 0.2157 (0.0144) && 22.5 (0.5) & 324.5 (10.5)  \\      
\bottomrule
\end{tabular}\\[1ex]
\end{table*}

\begin{table*}[htbp]
\centering
\caption{Comparison of training efficiency across DeepCDE, GCDS, and DDPM. Average (standard deviation) epoch time is reported in CPU seconds.}
\label{tab:epoch_time}
	\centering
	\footnotesize
	\setlength{\tabcolsep}{4pt}
	\begin{tabular}{@{}llccllc@{}}
	\toprule
	\textbf{Model} & \textbf{Method} & \textbf{Epoch Time} & & \textbf{Model} & \textbf{Method} & \textbf{Epoch Time}\\
	\midrule
	M1      & DeepCDE   & 0.2199 (0.1300)  && M6 & DeepCDE   & 0.2580 (0.2973) \\
        	& GCDS      & 0.8235 (1.2695)  &&    & GCDS      & 0.5647 (0.2618) \\
        	& DDPM      & 0.3657 (0.0293)  &&    & DDPM      & 0.3674 (0.0399) \\
	\midrule
	M2      & DeepCDE   & 0.2199 (0.1941)  && M7  & DeepCDE   &  0.2373 (0.1728) \\
        	& GCDS      & 2.4470 (4.6249)  &&     & GCDS      &  0.6129 (0.6331) \\
        	& DDPM      & 0.3688 (0.0335)  &&     & DDPM      &  0.3774 (0.0918)  \\	   

	\midrule
	M3      & DeepCDE   & 0.1981 (0.1175)  && M8   & DeepCDE   & 0.1397 (0.0091)\\
            & GCDS      & 0.6978 (0.9915)  &&      & GCDS      & 0.6756 (0.8536) \\
        	& DDPM      & 0.3651 (0.0343)  &&      & DDPM      &  0.3661 (0.0341) \\ 	 
	\midrule
	M4     & DeepCDE   & 0.1369 (0.0100)   && M9 & DeepCDE    & 0.1688 (0.0669) \\
            & GCDS      & 0.5366 (0.0273)  &&    & GCDS       & 0.5414 (0.0381) \\ 
        	  & DDPM      & 0.3584 (0.0420)  &&    & DDPM       & 0.3636 (0.0365)  \\       
      
	\midrule   
	M5      & DeepCDE   & 0.1962 (0.1112)  &&  M10 & DeepCDE   & -- \\
        	& GCDS      & 1.0545 (0.8006)  &&      & GCDS      & 0.5580 (0.0284) \\     
        	& DDPM      & 0.3679 (0.0438)  &&      & DDPM      & 0.3800 (0.0327)  \\ 	 
            
	\bottomrule
	\end{tabular}\\[1ex]
\vspace{0.5ex}
\end{table*}

\section{Discussion}\label{s:discussion}
Conditional distribution estimation is a central theme in modern statistics, and the landscape of available methodologies is remarkably diverse. This article has surveyed four representative approaches: single-index dimension reduction, basis-expansion regression, and two classes of generative simulation-based methods, and evaluated their performance across various scenarios in which the predictor dimension ranges from low to moderately high.
The main empirical findings from our simulation study can be summarized as follows.

For conditional distribution estimation, DDPM, an autoregressive generative model, appears to be the most robust approach among the five methods considered based on the simulation study.

FlexCode and DeepCDE, which are nonparametric estimation approaches using basis expansions, offer a certain degree of flexibility for conditional density estimation when an appropriate basis type is chosen and a sufficiently large number of basis functions are used. However, their performance may still be less competitive relative to DDPM, which employs sufficiently expressive model architectures.

FlexCode estimates the basis coefficient functions using nonparametric regression methods; hence, it can perform poorly when the underlying scenario is unfavorable for the chosen nonparametric regression approach, such as in the presence of heteroscedastic noise. In such cases, using DeepCDE may lead to improved conditional distribution estimation.

DeepCDE, GCDS, and DDPM all employ neural networks
and therefore require proper hyperparameter tuning, which can be challenging in complex conditional distribution settings. In particular, GCDS relies on a min-max distribution-matching objective and is often sensitive to hyperparameter choices and prone to training instability, based on our empirical experiences.

The classical dimension-reduction method for conditional distribution estimation proposed by Hall and Yao is generally less competitive than the other four approaches considered in our study.

\normalem
\bibliographystyle{apalike}   
\bibliography{Generative}

@article{han2022card,
  title={Card: Classification and regression diffusion models},
  author={Han, Xizewen and Zheng, Huangjie and Zhou, Mingyuan},
  journal={Advances in Neural Information Processing Systems},
  volume={35},
  pages={18100--18115},
  year={2022}
}

@article{dhariwal2021diffusion,
  title={Diffusion models beat gans on image synthesis},
  author={Dhariwal, Prafulla and Nichol, Alexander},
  journal={Advances in neural information processing systems},
  volume={34},
  pages={8780--8794},
  year={2021}
}

@article{fu2024unveil,
  title={Unveil conditional diffusion models with classifier-free guidance: A sharp statistical theory},
  author={Fu, Hengyu and Yang, Zhuoran and Wang, Mengdi and Chen, Minshuo},
  journal={arXiv preprint arXiv:2403.11968},
  year={2024}
}

@article{goodfellow2014generative,
  title={Generative adversarial nets},
  author={Goodfellow, Ian J and Pouget-Abadie, Jean and Mirza, Mehdi and Xu, Bing and Warde-Farley, David and Ozair, Sherjil and Courville, Aaron and Bengio, Yoshua},
  journal={Advances in neural information processing systems},
  volume={27},
  year={2014}
}

@article{henzi2023distributional,
  title={Distributional (single) index models},
  author={Henzi, Alexander and Kleger, Gian-Reto and Ziegel, Johanna F},
  journal={Journal of the American Statistical Association},
  volume={118},
  number={541},
  pages={489--503},
  year={2023},
  publisher={Taylor \& Francis}
}

@article{ramdas2017wasserstein,
  title={On {W}asserstein two-sample testing and related families of nonparametric tests},
  author={Ramdas, Aaditya and Garc{\'\i}a Trillos, Nicol{\'a}s and Cuturi, Marco},
  journal={Entropy},
  volume={19},
  number={2},
  pages={47},
  year={2017},
  publisher={MDPI}
}

@article{dalmasso2020conditional,
  title={Conditional density estimation tools in python and R with applications to photometric redshifts and likelihood-free cosmological inference},
  author={Dalmasso, Niccol{\`o} and Pospisil, Taylor and Lee, Ann B and Izbicki, Rafael and Freeman, Peter E and Malz, Alex I},
  journal={Astronomy and Computing},
  volume={30},
  pages={100362},
  year={2020},
  publisher={Elsevier}
}

@article{spokoiny2013local,
  title={Local quantile regression},
  author={Spokoiny, Vladimir and Wang, Weining and H{\"a}rdle, Wolfgang Karl},
  journal={Journal of Statistical Planning and Inference},
  volume={143},
  number={7},
  pages={1109--1129},
  year={2013},
  publisher={Elsevier}
}

@article {Austin15,
    AUTHOR = {Austin, Tim},
     TITLE = {Exchangeable random measures},
   JOURNAL = {Ann. Inst. Henri Poincar\'e{} Probab. Stat.},
  FJOURNAL = {Annales de l'Institut Henri Poincar\'e{} Probabilit\'es et
              Statistiques},
    VOLUME = {51},
      YEAR = {2015},
    NUMBER = {3},
     PAGES = {842--861},
      ISSN = {0246-0203,1778-7017},
   MRCLASS = {60G09 (60G57)},
  MRNUMBER = {3365963},
MRREVIEWER = {Flora\ Koukiou},
       DOI = {10.1214/13-AIHP584},
       URL = {https://doi.org/10.1214/13-AIHP584},
}

@inproceedings{yang2025conditional,
  title={Conditional diffusion models based conditional independence testing},
  author={Yang, Yanfeng and Li, Shuai and Zhang, Yingjie and Sun, Zhuoran and Shu, Hai and Chen, Ziqi and Zhang, Renming},
  booktitle={Proceedings of the AAAI Conference on Artificial Intelligence},
  volume={39},
  number={21},
  pages={22020--22028},
  year={2025}
}

@article{izbicki2017converting,
  title={Converting high-dimensional regression to high-dimensional conditional density estimation},
  author={Izbicki, Rafael and Lee, Ann B.},
journal={Electron. J. Statist.},
  year={2017},
volume={11(2)},
  pages={2800-2831}
}

@article{yu1998local,
  title={Local linear quantile regression},
  author={Yu, Keming and Jones, MC1614628},
  journal={Journal of the American statistical Association},
  volume={93},
  number={441},
  pages={228--237},
  year={1998},
  publisher={Taylor \& Francis}
}

@article{hall2005approximating,
  title={Approximating conditional distribution functions using dimension reduction},
  author={Peter Hall and Qiwei Yao},
  journal={Annals of Statistics},
  year={2005},
  volume={33},
  pages={1404-1421}
}

@article{zhou2023deep,
  title={A deep generative approach to conditional sampling},
  author={Zhou, Xingyu and Jiao, Yuling and Liu, Jin and Huang, Jian},
  journal={Journal of the American Statistical Association},
  volume={118},
  number={543},
  pages={1837--1848},
  year={2023},
  publisher={Taylor \& Francis}
}

@article{hall1999methods,
  title={Methods for estimating a conditional distribution function},
  author={Hall, Peter and Wolff, Rodney CL and Yao, Qiwei},
  journal={Journal of the American Statistical association},
  volume={94},
  number={445},
  pages={154--163},
  year={1999},
  publisher={Taylor \& Francis}
}

@article{ho2020denoising,
  title={Denoising diffusion probabilistic models},
  author={Ho, Jonathan and Jain, Ajay and Abbeel, Pieter},
  journal={Advances in neural information processing systems},
  volume={33},
  pages={6840--6851},
  year={2020}
}

@article{schmidt2020nonparametric,
author = {Johannes Schmidt-Hieber},
title = {{Nonparametric regression using deep neural networks with ReLU activation function}},
volume = {48},
number = {4},
journal = {The Annals of Statistics},
publisher = {Institute of Mathematical Statistics},
pages = {1875-1897},
year = {2020},
doi = {10.1214/19-AOS1875},
URL = {https://doi.org/10.1214/19-AOS1875}
}

@article{bauer2019,
  title={On deep learning as a remedy for the curse of dimensionality in nonparametric regression},
  author={Bauer, Benedikt and K{\"o}hler, Michael},
  journal={The Annals of Statistics},
  volume={47},
  number={4},
  pages={2261--2285},
  year={2019},
  doi={10.1214/18-AOS1747},
  publisher={The Institute of Mathematical Statistics}
}

@book{Goodfellow-et-al-2016,
    title={Deep Learning},
    author={Ian Goodfellow and Yoshua Bengio and Aaron Courville},
    publisher={MIT Press},
    note={\url{http://www.deeplearningbook.org}},
    year={2016}
}

@article{nguyen2010estimating,
  title={Estimating divergence functionals and the likelihood ratio by convex risk minimization},
  author={Nguyen, XuanLong and Wainwright, Martin J and Jordan, Michael I},
  journal={IEEE Transactions on Information Theory},
  volume={56},
  number={11},
  pages={5847--5861},
  year={2010},
  publisher={IEEE}
}

@book{kallenberg2021foundations,
  title={Foundations of Modern Probability},
  author={Kallenberg, Olav},
  year={2021},
  publisher={Springer}
}

@article{hinton2006reducing,
  title={Reducing the dimensionality of data with neural networks},
  author={Hinton, Geoffrey E and Salakhutdinov, Ruslan R},
  journal={science},
  volume={313},
  number={5786},
  pages={504--507},
  year={2006},
  publisher={American Association for the Advancement of Science}
}

@article{song2020denoising,
  title={Denoising diffusion implicit models},
  author={Song, Jiaming and Meng, Chenlin and Ermon, Stefano},
  journal={Proceedings of the International Conference on Learning Representations},
  year={2021},
  note = {Available at arXiv:2010.02502}
}

@article{flamary2021pot,
  title={Pot: Python optimal transport},
  author={Flamary, R{\'e}mi and Courty, Nicolas and Gramfort, Alexandre and Alaya, Mokhtar Z and Boisbunon, Aur{\'e}lie and Chambon, Stanislas and Chapel, Laetitia and Corenflos, Adrien and Fatras, Kilian and Fournier, Nemo and others},
  journal={Journal of Machine Learning Research},
  volume={22},
  number={78},
  pages={1--8},
  year={2021}
}

@article{bonneel2015sliced,
  title={Sliced and {R}adon {W}asserstein barycenters of measures},
  author={Bonneel, Nicolas and Rabin, Julien and Peyr{\'e}, Gabriel and Pfister, Hanspeter},
  journal={Journal of Mathematical Imaging and Vision},
  volume={51},
  number={1},
  pages={22--45},
  year={2015},
  publisher={Springer}
}

\clearpage
\onecolumn

\clearpage

\end{document}